\documentclass[journal]{IEEEtran}

\pdfoutput=1

\usepackage{cite} 
\usepackage{amsmath,graphicx} 
\usepackage{amsfonts,amssymb}
\usepackage{algorithmic}
\usepackage{algorithm}

\usepackage{hyperref}
\usepackage{url}

\begin{document}

% Define document title and author
\title{Fast and Robust Multiple ColorChecker Detection using Deep Convolutional Neural Networks}

\author{Pedro D. Marrero Fernandez, Fidel A. Guerrero-Pe\~{n}a, Tsang Ing Ren,~\IEEEmembership{Member,~IEEE,} and Jorge J. G. Leandro % <-this % stops a space

\thanks{Pedro D. Marrero Fernandez, Fidel A. Guerrero-Pe\~{n}a and Tsang Ing Ren are with Centro de Inform\'{a}tica, Universidade Federal de Pernambuco, Recife, Brazil (e-mail:pdmf@cin.ufpe.br; fagp@cin.ufpe.br; tir@cin.ufpe.br).}% <-this % stops a space
\thanks{Jorge J. G. Leandro is with Motorola Mobility, LLC, Brazil (e-mail:jleandro@motorola.com )}}% <-this % stops a space

%\thanks{Thanks all}}
%\markboth{XXXX}{}

\maketitle

% Write abstract here
\begin{abstract}
ColorCheckers are reference standards that professional photographers and filmmakers use to ensure predictable results under every lighting condition. The objective of this work is to propose a new fast and robust method for automatic ColorChecker detection. The process is divided into two steps: (1) ColorCheckers localization and (2) ColorChecker patches recognition. For the ColorChecker localization, we trained a detection convolutional neural network using synthetic images. The synthetic images are created with the 3D models of the ColorChecker and different background images. The output of the neural networks are the bounding box of each possible ColorChecker candidates in the input image. Each bounding box defines a cropped image which is evaluated by a recognition system, and each image is canonized with regards to color and dimensions. Subsequently, all possible color patches are extracted and grouped with respect to the center's distance. Each group is evaluated as a candidate for a ColorChecker part, and its position in the scene is estimated. Finally, a cost function is applied to evaluate the accuracy of the estimation. The method is tested using real and synthetic images. The proposed method is fast, robust to overlaps and invariant to affine projections. The algorithm also performs well in case of multiple ColorCheckers detection.
\end{abstract}


\section{Introduction}
\label{intro}

The illumination of a scene highly influences the reproduction of the colors in images captured with digital cameras. Given a camera sensor with a Spectral Sensitivity, color renderings can deviate significantly from colors perceived by human eyes depending on the Color Stimulus, that is, the product between the Spectral Power Distribution of the incoming light and the Spectral Reflectance of the object. However, accurate color reproduction still requires colorimetric camera calibration for different illuminations that is usually done using a ColorChecker (CC) that shows predefined regions with specified colors. Manufacturers offer various checker models for specific applications \cite{Ernst2013CheckCalibration}.

ColorChecker Targets are reference standards that professional photographers and filmmakers use to ensure predictable results under every lighting condition. The use of the CC speeds up the color adjustment in the process of obtaining accurate colors. Therefore, minimizing tedious work of trial and error color adjustments while editing or color grading. In order to scale the assessment of the color accuracy and auto white balance for a large number of images, it is necessary to automate this process, thus automating CC detection is also required.

\begin{figure}[t]
\centering
\begin{tabular}{cc}
\includegraphics[width=1.5in]{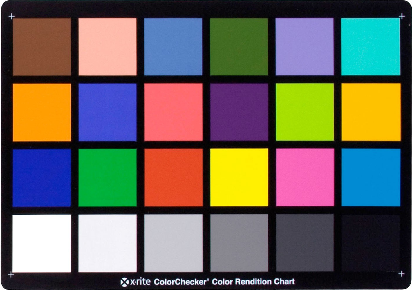} &
\includegraphics[width=1.5in]{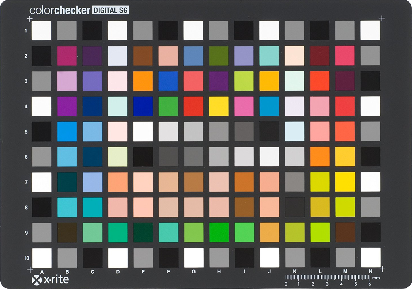} \\
a & b 
\end{tabular}
\caption{Examples of the X-Rite ColorCheckers: a) ColorChecker\textregistered~Classic; b) ColorChecker \textregistered~Digital SG.}
\label{fig:colorcheckers}
%\vspace{-0.5cm}
\end{figure}

Currently, there are several published papers in the literature on geometric camera calibration. However, only a few publications on the detection of ColorChecker in images. Many types of ColorCheckers exist, each one specifically designed for a different class of device and for the property to assess. \figurename~\ref{fig:colorcheckers} shows two of the most used ColorChecker's: X-Rite ColorChecker\textregistered~Classic (CCC) and X-Rite ColorChecker\textregistered~Digital SG (CSG). The CCC is an 8 $\times$ 11 inch chart which consists of 24 patches with 18 familiar colors and six grayscale levels having optical densities from 0.05 to 1.50 and a range of 4.8 f-stops. The colors are not highly saturated, and the ColorChecker quality is very high.  Each patch is printed separately using controlled pigments, and the patches have a smooth matte surface. The CCC is a standard color target. The exact description of these ColorCheckers can be found on the X-Rite website\footnote{\url{http://xritephoto.com/colorchecker-targets}}.

Tajbakhsh and Grigat proposed an algorithm for semi-automatic CCC detection in distorted images \cite{Tajbakhsh2008SemiautomaticImages}. Initially, the user selects four corners in the ColorChecker image, and the system estimates the position of all color patches using projective geometry. The image is processed with a Sobel kernel, a morphological operator, and a threshold is applied converting it into a binary image and the connected patches are found. 

Kapusi et al. proposed a method combining geometric and colorimetric camera calibration with a unified calibration checker \cite{kapusi2010simultaneous}. A black and white chessboard pattern is used for the geometric calibration, and 24 circular reference color patches are placed in squares center for the color calibration. The chessboard pattern detection is obtained using the OpenCV library \cite{Bradski200807Library}. A subsequent region growing algorithm segments the circular color regions \cite{kapusi2010simultaneous}.

Bianco et al. \cite{Bianco2011} presented a method for single CCC detection. The method locates the ColorChecker by finding the coordinates on the local color descriptors spaces using the SIFT descriptors and recognizes the ColorChecker using an optimization function. 

Ernst et al. proposed a robust algorithm to detect and track color calibration checkers in images  \cite{Ernst2013CheckCalibration}. The automatic CCC detection procedure uses a cost function to find the checker in an image. The cost function compares the colors of the patches with the reference colors and the standard deviations of the colors within the patch regions. Four corners of the checker model are projected with the use of the Direct Linear Transformation to find the coordinates in an image. The coordinates of the CCC are obtained by minimizing the cost function using the Levenberg-Marquardt algorithm. The procedure can detect ColorChecker if it is in front of the camera within an allowed operating range. 

Andrzej et al. present an algorithm for automatic ColorChecker detection and color patch value extraction. The algorithm can detect different types of CC in various images. This method performs a k-means kind of clustering over the RGB color space, using 25 colors as centroids (24 color chart + 1 background), which generates a segmentation of the regions corresponding to the patches. In the sequel, it eliminates some of the regions according to a criterion of shape and area. It then groups the regions by area and by the distance between centers of the patches. Finally, it estimates the bounding parallelogram (hypothesis) on the convex hull of the obtained groups \cite{schwarz1995minimal}. This method assumes that the final hypothesis is a parallelogram, which is not necessarily true \cite{Kordecki2014}.

Software tools to detect CCC, such as CCFind\footnote{\url{http://issl.udayton.edu/index.php/research/ccfind/}} and MacDuff\footnote{\url{https://github.com/ryanfb/macduff}}, are freely available. CCFind is implemented in Matlab (Mathworks Inc.) and returns the coordinates of the center points of the color patches. By not using colors as a cue, it can be used with unconventional lighting and multispectral sensors. On the other hand, MacDuff is implemented in C++ and uses OpenCV library. By performing geometrical operations (rotations, scaling, etc.) and by computing a distance metric between the colors, the ColorChecker is finally found.

Some commercial software is also capable of doing a semi-automatic ColorChecker target detection. Examples are the X-Rite ColorChecker Passport Camera Calibration Software\footnote{\url{http://www.xrite.com/}}, Imatest\footnote{\url{http://www.imatest.com/}} and BabelColor PatchTool\footnote{\url{http://www.babelcolor.com/}}, which tries to perform an initial automatic detection. These softwares however usually rely on human intervention to manually mark or correct the detected reference target (by dragging the cursor), after which color correction is performed. Since manual intervention is not practical in mass digitalization processes for obvious reasons of cost and speed, it is interesting to develop a fully automatic tool for the detection of one or several ColorChecker targets in digital images.

\begin{figure*}[ht]
\centering
\begin{tabular}{cccc}
\includegraphics[width=1.6in]{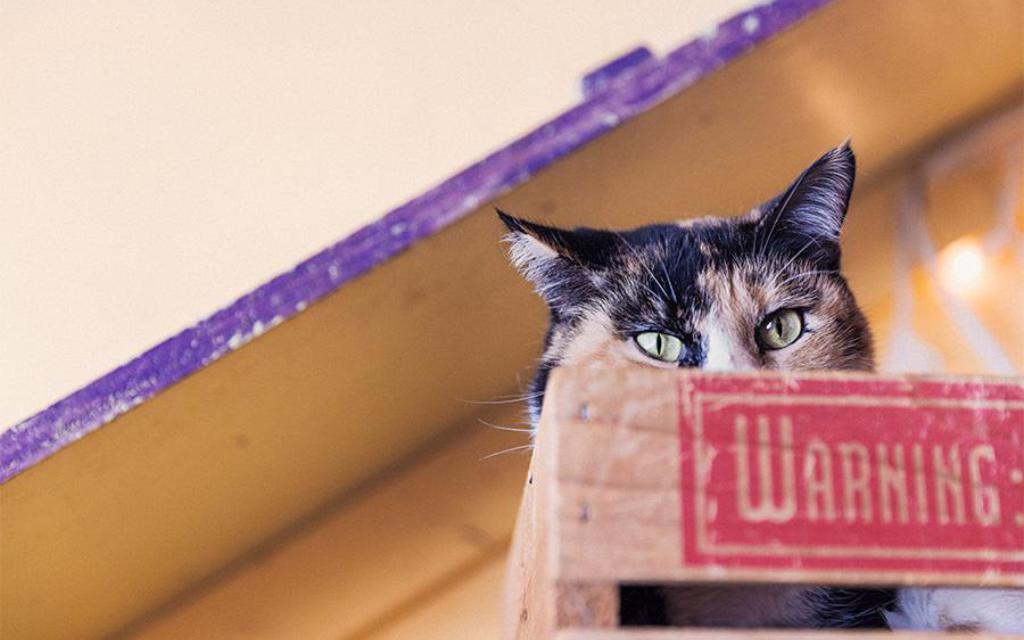} &
\includegraphics[width=1.6in]{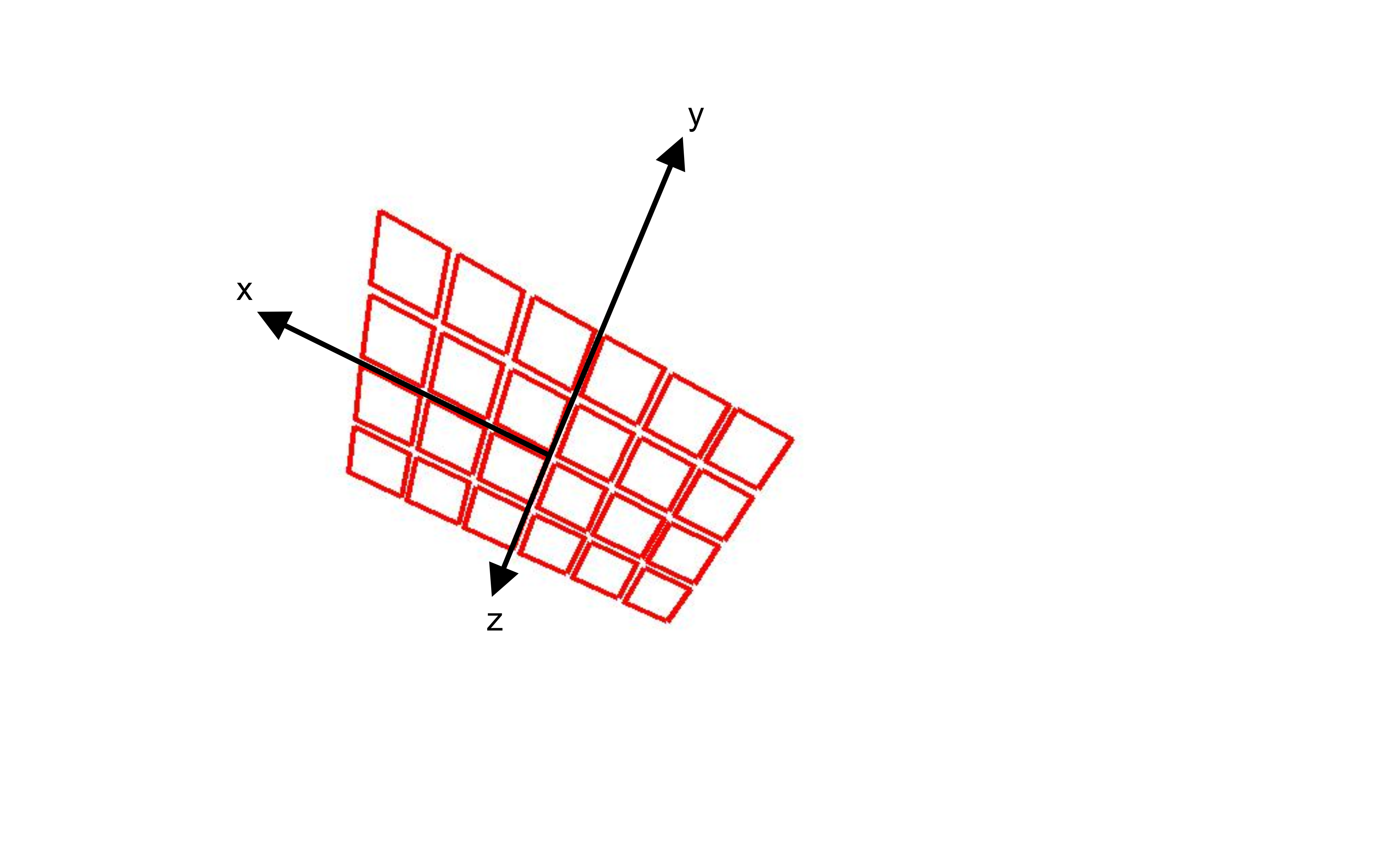} &
\includegraphics[width=1.6in]{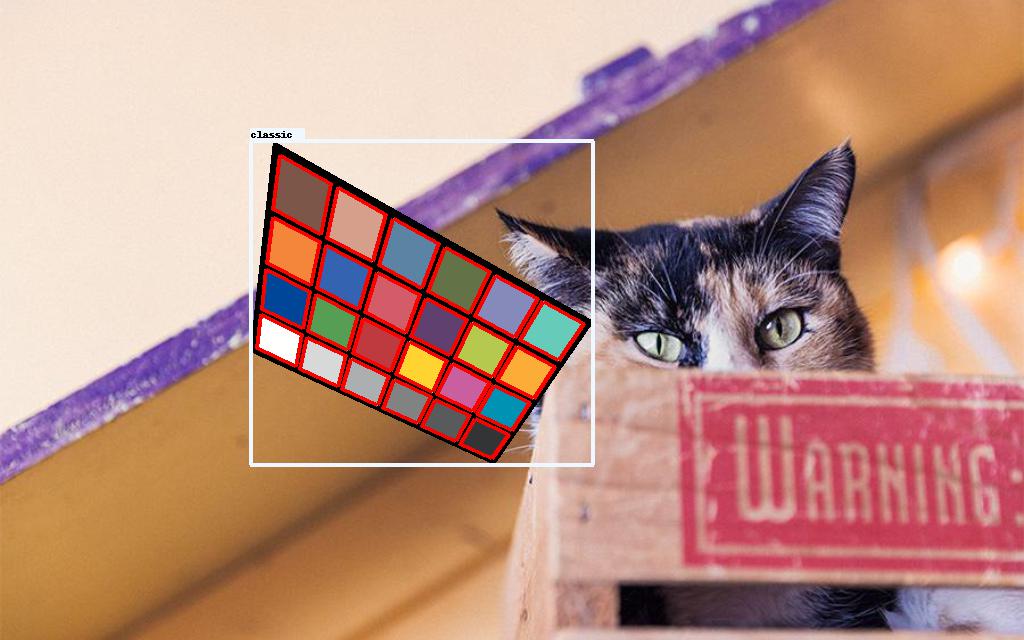} &
\includegraphics[width=1.6in]{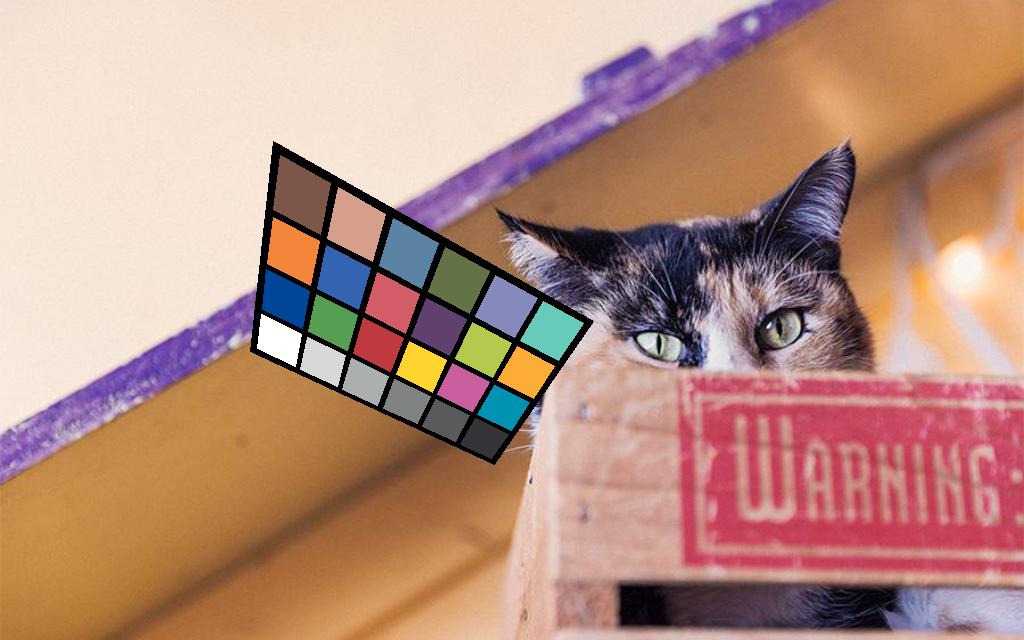} \\
a & b & c & d
\end{tabular}
\caption{The synthetic dataset generation process: from a) Original initial image, b) generated 3d model, c) random placement of the ColorChecker and d) final synthetic image.}
\label{fig:synthetic_dataset_create}
%\vspace{-0.5cm}
\end{figure*}

The objective here is to propose a fast and robust method for automatic multiple ColorChecker's detections in the image. We applied a convolutional neural network for the localization of the CCC trained over a synthetic dataset (the generation process of this dataset is shown in \figurename~\ref{fig:synthetic_dataset_create})  and a new ColorChecker patch recognition method.

\section{Proposed Method}

\begin{figure*}[t!]
\centering
\includegraphics[width=7.0in]{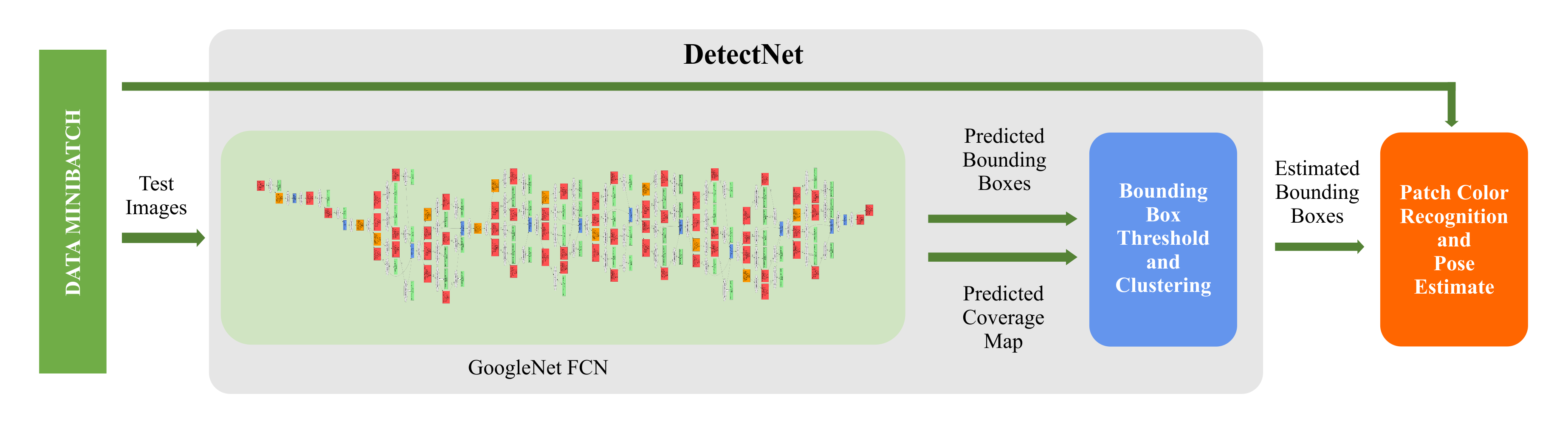}
\caption{The overview of the complete proposed system showing its different components.  }
\label{fig:pipeline_general}
\end{figure*}

We present a novel methodology for the detection and recognition of multiple ColorChecker's Classic. This algorithm is easily adaptable to any ColorChecker that has a set of uniformly colored patches and a parallelogram shape. \figurename~\ref{fig:pipeline_general} shows an overview of our system pipeline. The method is comprised of two primary stages: (1) CCC candidates location in the image and (2) CCC color patches recognition and pose estimation. The code for the ColorChecker-detection\footnote{\url{https://github.com/pedrodiamel/colorchacker-detection}} was made available in a repository.

Stage (1) is responsible for the localization of all the CCC in the image. Once the CCCs are localized, the stage (2), focuses on the image regions analysis with a high probability of containing a CCC. The initial localization in stage (1) increases the speed of the system and its accuracy as will be shown in the experiment section. Each component of the pipeline is described in this section.

\subsection{Deep Learning ColorChecker Localization}

The first step of the proposed method is the localization of all possible ColorChecker candidates. In \cite{Bianco2011} the SIFT descriptor is employed for this task. However, this class of methods is not robust to changes in the illumination (over or underexposed) and has problems with scalability for multiple ColorChecker's in the image. Moreover, SIFT is a patented invention. A Convolutional Neural Network model provides an end-to-end solution suited for this problem.

In recent years, convolutional neural networks (CNN) with deep architectures have outperformed traditional machine learning approaches for several computer vision tasks, where a large amount of labeled training data is available. Generally, the harder the task, the deeper the needed neural network, and more training data is also required \cite{Le2017UsingReasoning}.

When labeled training data is unavailable, synthetic data can be generated to compensate for this lack of data. Unsupervised generative model is a recent promising approach but generally, has a slow test-time inference, because it needs new data. Remarkable results have been obtained using synthetic data solutions, including the limited form of data augmentation \cite{Simard2003BestAnalysis,krizhevsky2012imagenet}. 

An interesting work on text recognition in the wild \cite{Jaderberg2014SyntheticRecognition,Jaderberg2016ReadingNetworks,Gupta2016SyntheticImages} is an example, which was achieved by training a neural network to recognize text using synthetically generated realistic renders. Goodfellow et al. \cite{Goodfellow2014Multi-digitNetworks} addressed the problem of the recognition of house numbers in images from the Google Street View in a supervised fashion, also solving reCaptcha \cite{Ahn2008ReCAPTCHA:Measures} images using synthetic data to train a recognition neural network from image to latent text. They were able to access the actual reCaptcha generative model. Therefore, they could generate millions of labeled instances to use in a standard supervised learning pipeline.  More recently, Stark et al. \cite{Stark2015CAPTCHALearning} also used synthetic data for captcha solving and Wang et al. \cite{Wang2015DeepFont:Image} for font identification.

\begin{figure}[!t]
\centering
\includegraphics[width=2.7in]{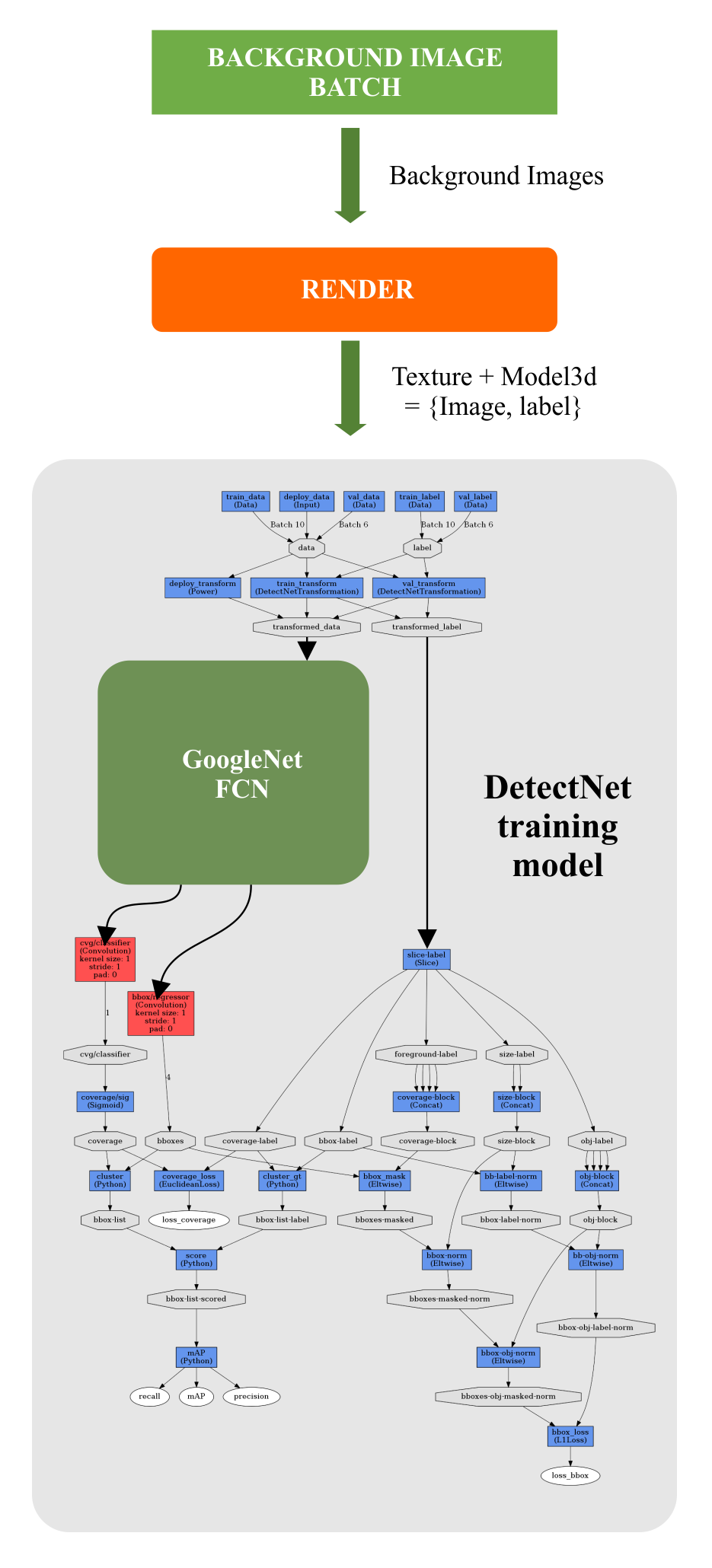}
\caption{The proposed DetectNet neural network training model with a new layer (Render). The CNN training is performed using a synthetic data.}
\label{fig:detection_net_train}
%\vspace{-0.5cm}
\end{figure}

Le et al. \cite{Le2017UsingReasoning} demonstrated that the use of synthetic data to train a neural network is equivalent to train an artifact to do amortized approximate inference \cite{Gershman2014AmortizedReasoning}. In this work, we created a new layer to generate synthetic data. \figurename~\ref{fig:detection_net_train} shows the training scheme for the DetectNet\footnote{\url{https://github.com/NVIDIA/DIGITS/tree/master/examples/object-detection}} using synthetic data. We describe the generation model similar to that by Le et al. \cite{Le2017UsingReasoning}.

The proposed synthetic data generation model for ColorChecker localization specifies the joint densities $p(x, y)$, that defines the latent random variable $x$ and the corresponding ColorChecker image $y$. The latent structured random variable $x = \{C, \epsilon_{1:C}, i_{1:C}\}$ includes $C$, the number of ColorCheckers in the image, $\epsilon_{1:C}$, a multidimensional structured parameter set that controls the CCC-rendering such as the various deformations types (affine deformations, color deformations, etc), and $i_{1:C}$, the ColorChecker identities. We use a custom stochastic CC renderer $R$ to generate each image $y$. The synthetic data generator corresponds to the model:

\begin{equation}
 y|x \sim R(x)
\end{equation}

In particular, the model places uniform distributions over different intervals for C, $\epsilon_{1:C}$, and $i_{1:C}$, thus generating the synthetic training data $\{(x^n , y^n )\}$, where $n$ is the total number of images.  

\begin{figure}[!t]
\centering
\begin{tabular}{ccc}
\includegraphics[width=1.5in]{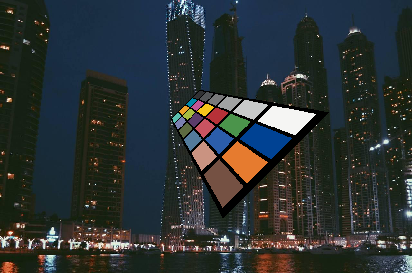} &
\includegraphics[width=1.5in]{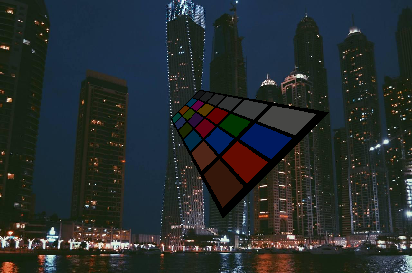} \\
a & b \\
\end{tabular}
\caption{Example of a) synthetic image without luminance adjustment (original) and b) image with luminance adjustment.}
\label{fig:ligth_render}
%\vspace{-0.5cm}
\end{figure}

The renderer $R$ adjusts the illumination of each ColorChecker so that it is inserted in the scene more realistically. To yield a more realistic insertion of the CCC in the scene, an additional step would be necessary to generate an alpha matte so that a final composite image of CCC and background could be constructed. The luminance channel of the ColorChecker model $I_{cc}$ is adjusted by multiplying it by the factor $\frac{I_{r}}{I_{cc}}$ where $I_{r}$ is the luminance of the region that contains it in the original image. \figurename~\ref{fig:ligth_render} shows the difference between the adjusted luminance (b) and the non-adjusted luminance.

For the detection model, we applied the DetectNet. The Fully-Convolutional Network (FCN) sub-network of DetectNet has the same structure as the GoogLeNet without the data input layers, final pooling layer and output layers \cite{szegedy2015going}. This has the benefit of allowing DetectNet to be initialized using a pre-trained GoogLeNet model, thereby reducing training time and improving final model accuracy. The fully connected layers predict the output probabilities and coordinates.

\subsection{Patch Color Recognition and Pose Estimation}

After the CC localization, the images are cropped using estimated bounding boxes, and the checker recognition step is applied to the cropped images. The proposed recognition method can be applied to the images without the previous CC detection step, but the performance is affected as shown in the results section.

\begin{figure}[!t]
\centering
\includegraphics[width=2.5in]{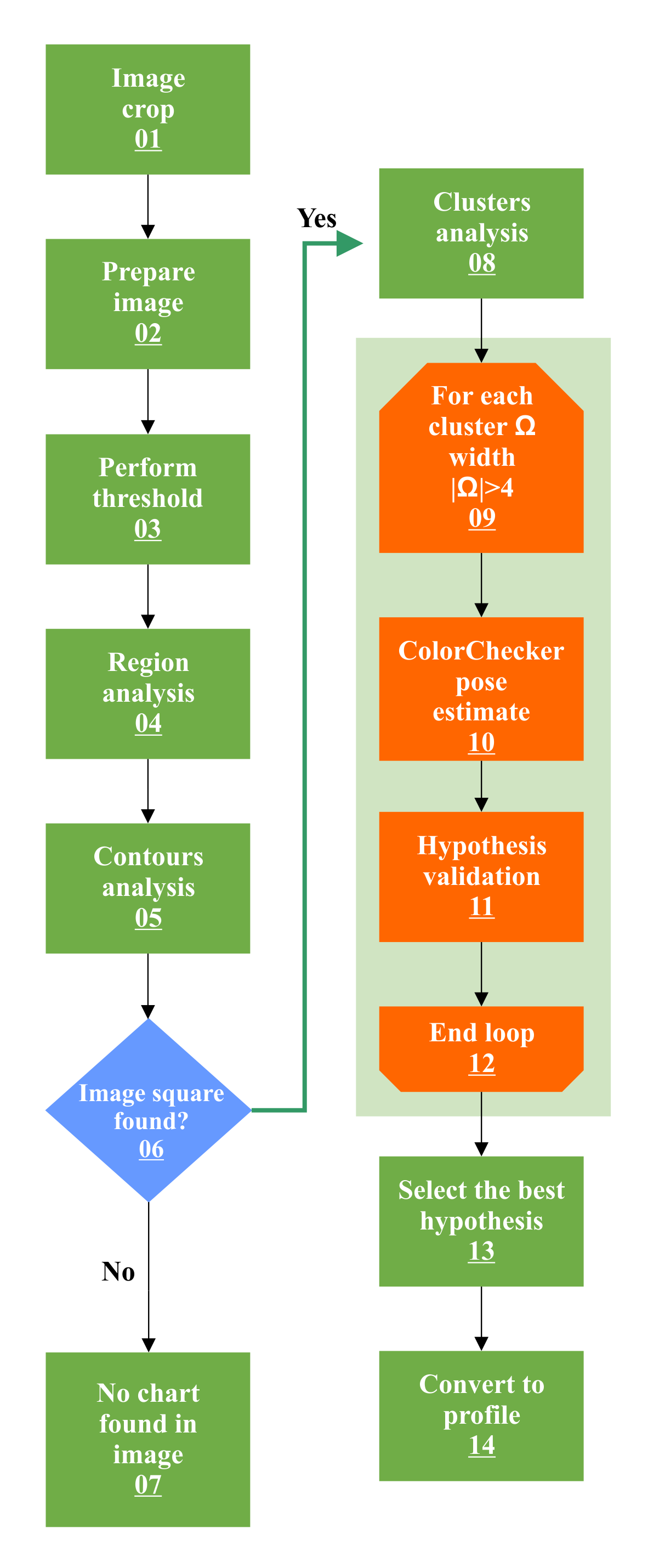}
\caption{The block diagram of the ColorChecker recognition system showing the 14 steps applied to obtain the CC estimated position.}
\label{fig:pipeline}
%\vspace{-0.5cm}
\end{figure}

\begin{figure}[!t]
\centering
\begin{tabular}{cc}
\includegraphics[width=1.5in]{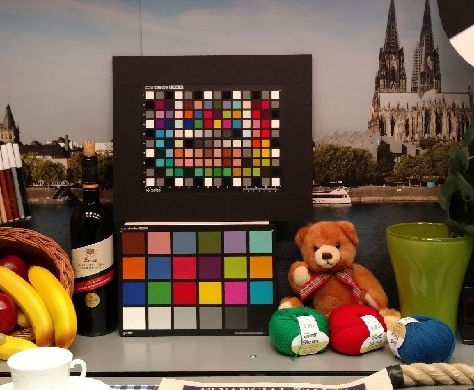} &
\includegraphics[width=1.5in]{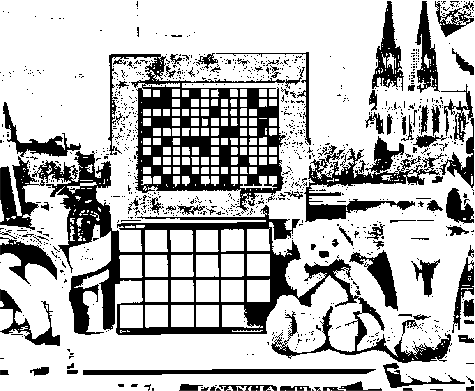} \\
01 & 03 \\
\includegraphics[width=1.5in]{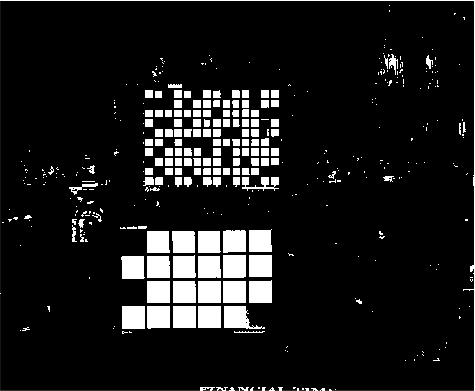} &
\includegraphics[width=1.5in]{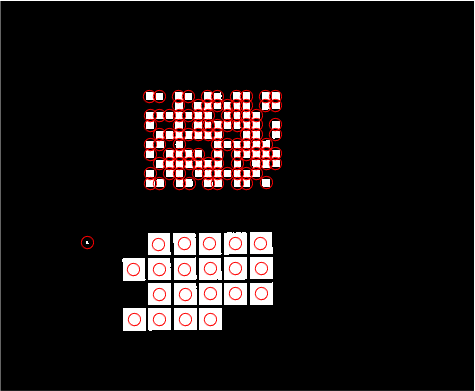} \\
04 & 05 \\
\includegraphics[width=1.5in]{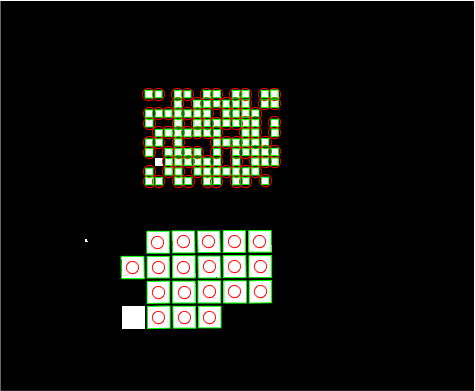} &
\includegraphics[width=1.5in]{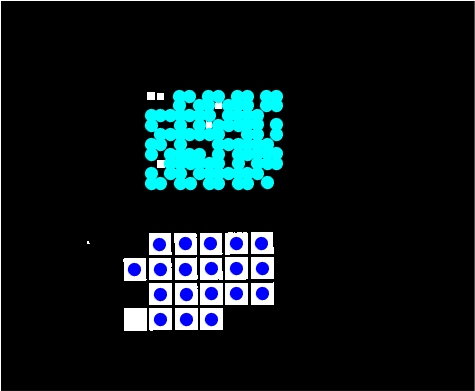} \\
06 & 08 \\
\end{tabular}
\caption{The output of the image for the steps 01, 03, 04, 05, 06 and 08 from the pipeline sequence for the ColorChecker recognition, as shown in \figurename~\ref{fig:pipeline}.}
\label{fig:sec_proc}
\end{figure}

\figurename~\ref{fig:pipeline} shows the complete recognition system pipeline. Each component of this pipeline is labeled with numbers from (01) to (14) and described in this section. The input of the system is an RGB image (01), as shown in \figurename~\ref{fig:sec_proc}. The input image is rescaled for that the smallest of the dimensions is equal to 400 and to keep the same aspect ratio. Also, we used Wiener filtering for noise suppression and RGB color normalization.

An adaptive thresholding (03) is applied to the canonical image, and morphological operations (04) techniques are applied to remove the undesired regions in the image border and isolated pixels that might appear due to the thresholding (see Figure 6 steps (03) and (04)). The candidate patches regions are analyzed and many false positive regions are discarded. The following features for each obtained region, which is a candidate patch, in the previous step are calculated (the best values for the parameters were selected after several trials):

\begin{itemize}
\item Convexity: The proportion of the number of pixels in the convex hull that are also in the region which is computed as $cc = Area/ConvexArea$. The selected threshold value was $cc > 0.90$.
\item Axes: An inscribed ellipse is determined for each region. Axes are defined as the relationship between the major and minor axis of this ellipse that has the same normalized second central moments as the region. This value is computed as $ac = AxisMin/AxisMax$ and the criterion of $ac > 0.4$ was applied.
\item Circularity: It measures the similarity between the shape of the region and a circumference. The circularity factor is computed as $cf = (4*\pi*Area) / Perimeter ^ 2$ and the range of $0.65 < cf < 0.97$ was applied.
\item Homogeneity: This feature corresponds to the entropy, which measures the level of information/disorganization present in the patch. Since color patches belonging to ColorCheckers are designed to be homogeneous, we define $hc = -\sum_i p(x_i)\log_2 p(x_i)$. The criterion of  $hc < 4.9$ was applied.
\end{itemize}

From the obtained region in the previous step, we are interested in the regions that are shaped as quadrilaterals (05). Quadrilaterals are the result of perspective deformations of the CC model. To determine these type of regions, the Ramer-Douglas-Peucker algorithm \cite{ramer72,peucker1973} is applied to approximate the region into a polygonal shape. Then, the polygons that contain only four points are selected. The minimal bounding parallelogram is applied to improve the results \cite{schwarz1995minimal}.

The second objective of this procedure is to assess all possible ColorChecker candidates in the scene (steps 08, 09, 10 and 11).  Because of that, a variant of the Hierarchical Compact Algorithm (HCA) \cite{gil2006general} is applied to the obtained groups of patches, as shown in \figurename~\ref{fig:sec_proc}, step (08). This method of clustering is based on the graph representation and the $B_0$-similarity\cite{gil2006general} concept for generating graphs. In this case, the graph vertices are the centers of the detected patches. A weight, based on the area ratio of the patches (defined in equation \ref{eq:bodist}), between each pair of nodes of the similarity graph is also established. The distance function is defined as:

\begin{equation} \label{eq:bodist} 
d_{ij} = w_{ij}*||X_i - X_j||_2
\end{equation}

\noindent
where $X_i$ is the $i$-th patch center, $w_{ij} = \frac{ |A_i - A_j|}{ (A_i + A_j)}$, and $A_i$ is the area corresponding to the $i$-th patch (node in the graph). For all pairs of vertices of the graph $\{X_i,X_j\}$, there exists an edge if $d_{ij} < B_{0i}$. A dynamic value for $B_{0i}$ was applied which is defined as:

\begin{equation} 
B_{0i} = AxisMax_{i}*1.65
\end{equation}

\begin{figure}[!t]
\centering
\includegraphics[width=2.5in]{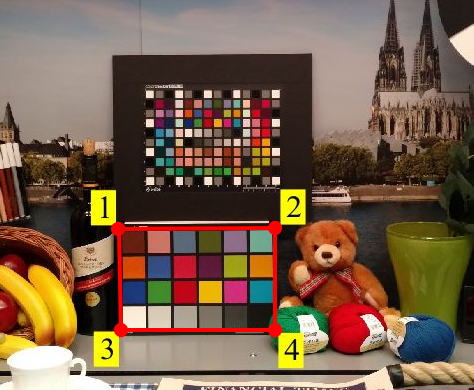}
\caption{The estimation of the ColorChecker Classic position and orientation, showing the four corners of the parallelogram.}
\label{fig:position_estimate}
\end{figure}

The groups of patches are formed by taking the connected components of the graph. Subsequently, each group is analyzed, and groups that have few elements (less than 4 in this case) are eliminated. In the next step, a CC estimated position is calculated (10). Also, the CC orientation in the image is identified, as shown in Fig \ref{fig:position_estimate}.

\begin{figure}[!t]
\centering
\includegraphics[width=1.25in]{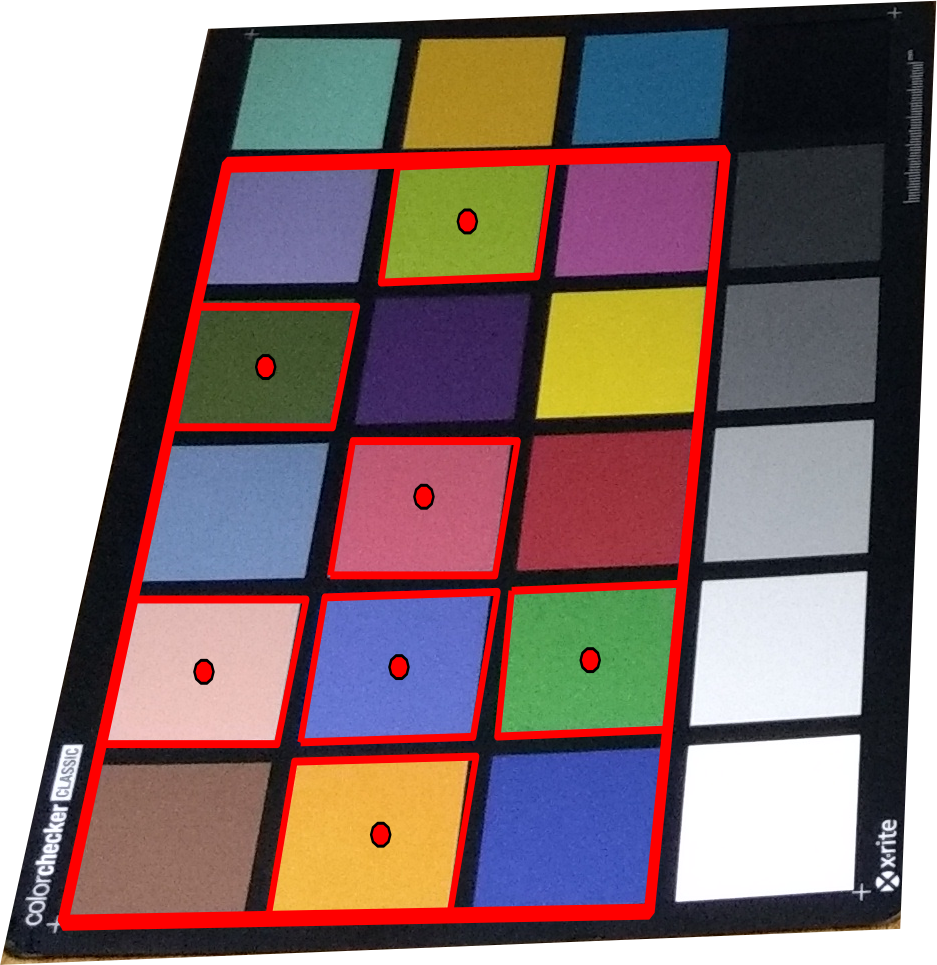}
\caption{The quadrilateral that best fits the ColorChecker is obtained applying the MEQ algorithm. The image shows an example of the output of Algorithm 1.}
\label{fig:minboundquadrilateral}
\end{figure}

\begin{figure}[!t]
\centering
\includegraphics[width=3.0in]{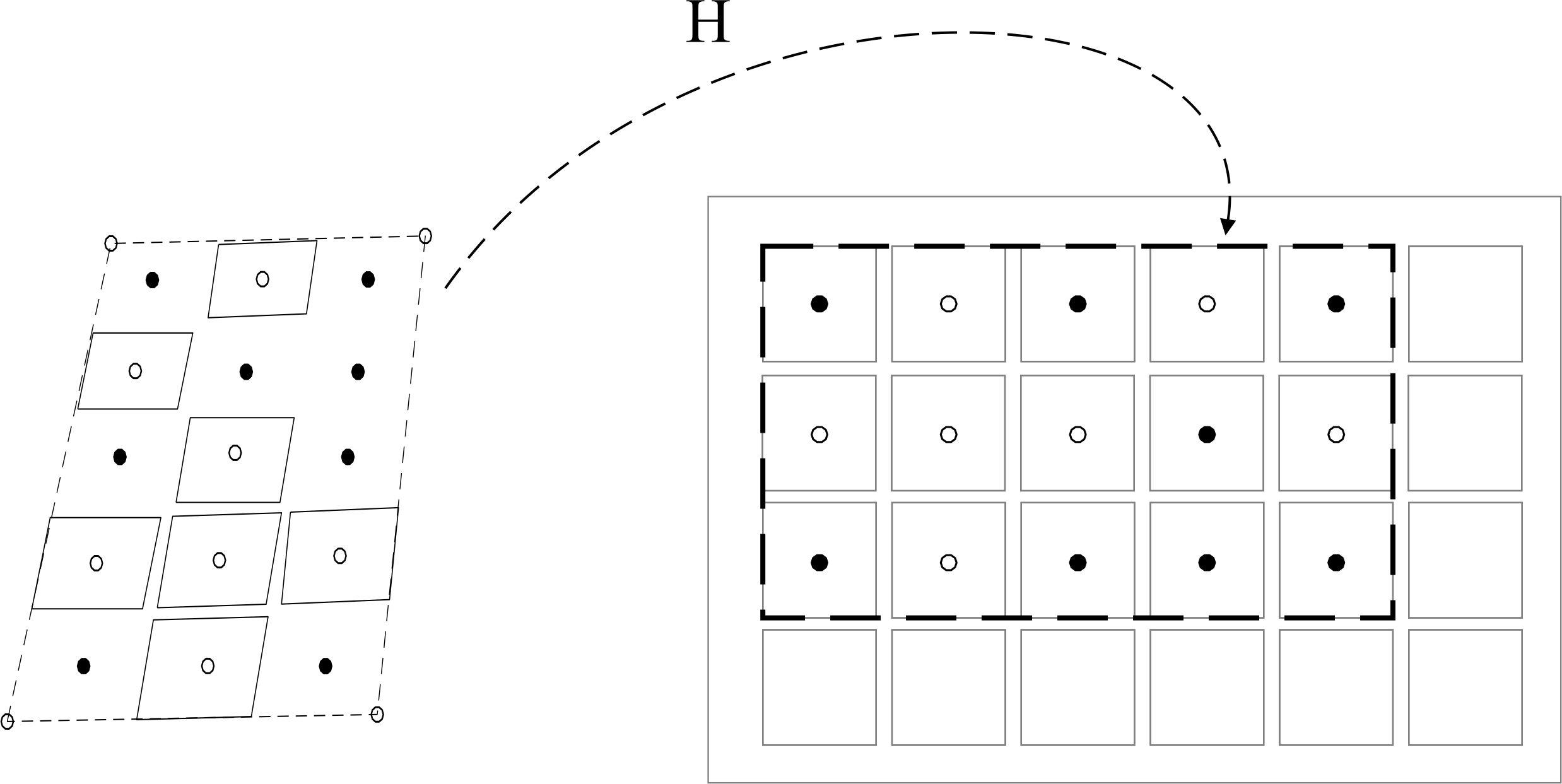}
\caption{The projection of the detected ColorChecker patches, showing the subset on the original CC that represents the obtained MEQ, the center position of the missing ColorChecker patches and the homography matrix H.}
\label{fig:homography}
\end{figure}

The quadrilateral that best fits the ColorChecker in that group is calculated for each obtained group. An example is shown in \figurename~\ref{fig:minboundquadrilateral} and the procedure is described in Algorithm 1. Unlike the method proposed by Schwarz et al. \cite{schwarz1995minimal}, this algorithm is more robust to affine transformations under several different conditions. Given the obtained minimum enclosing quadrilateral (MEQ), the center position of the missing patches and the homography matrix H are estimated with respect to the plane of the CC model, as shown in \figurename~\ref{fig:homography}.

\begin{figure}[!t]
\centering
\includegraphics[width=2.50in]{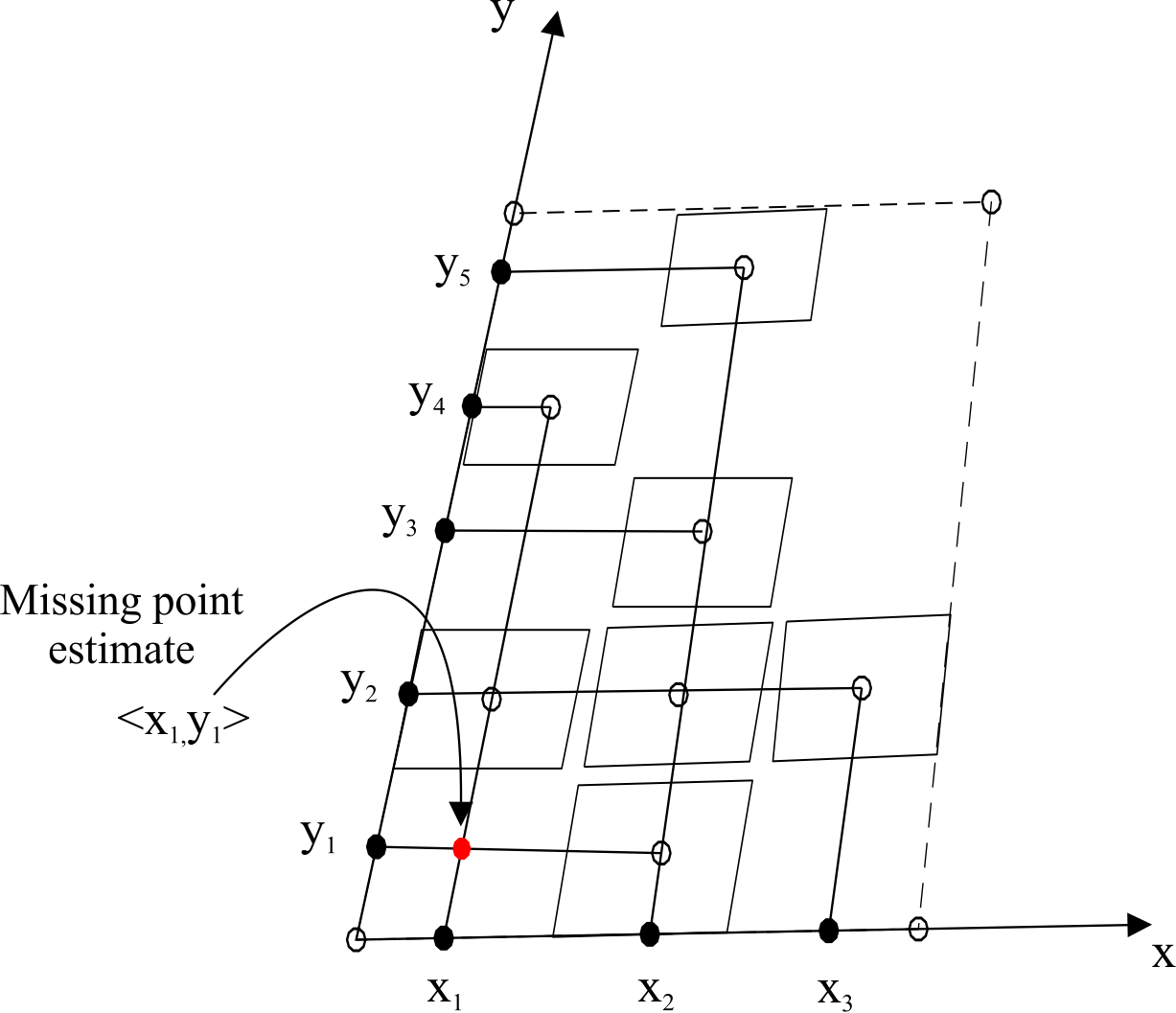}
\caption{Missing point estimate. The red point is an example of the estimation missing point.}
\label{fig:missingpoint}
\end{figure}

For missing point estimation, all centers points are projected into x and y axis of the estimated quadrilateral (see \figurename~\ref{fig:missingpoint}). The new points set is created with the combination of x and y projections. 

\begin{algorithm}[!t]
\caption{Minimum Enclosing Quadrilateral (MEQ).}
\label{alg:minboundquadrilateral}
\begin{algorithmic}[1]

\REQUIRE $Ch$ set of charts, where $ch_i \in Ch: ch_i=\{p_1,p_2,p_3,p_4\}$ is a sort clockwise set of 4 corner points.

\STATE Let $L$ be the set of lines formed by any two consecutive points on a chart (for a chart $ch_i=\{p_1,p_2,p_3,p_4\}$, $ch_i \in Ch$, only four straight lines can be obtained $l_{12},l_{23},l_{34},l_{41}$).
$$ L = \{ l_{ij} = p_i \times p_j, ~ i=1,2,3,4; ~j=2,3,4,1 \}, ~ \forall ch_i $$

\STATE The straight lines are sorted according to how far they are from the all points set $P$.
For each $l_{i} \in L$ with $l_{i} = [n_x,n_y,d]$ in homogeneous coordinates:

$$ds_i = \sum( P*[n_x, n_y]_i^T + d < 0 )$$

\STATE The first four lines were selected, such that:

$$\theta > 30^\circ $$ with $$cos(\theta) = \frac{l_{i} l_{i+1}}{ ||l_{i}|||| l_{i+1}|| }$$

\STATE The output $B=\{p_1,p_2,p_3,p_4\}$, represent the intersection of the selected lines in the previous step.

\end{algorithmic}
\end{algorithm}

%\newpage

\begin{figure}[!t]
\centering
\begin{tabular}{cc}
\includegraphics[width=1.30in]{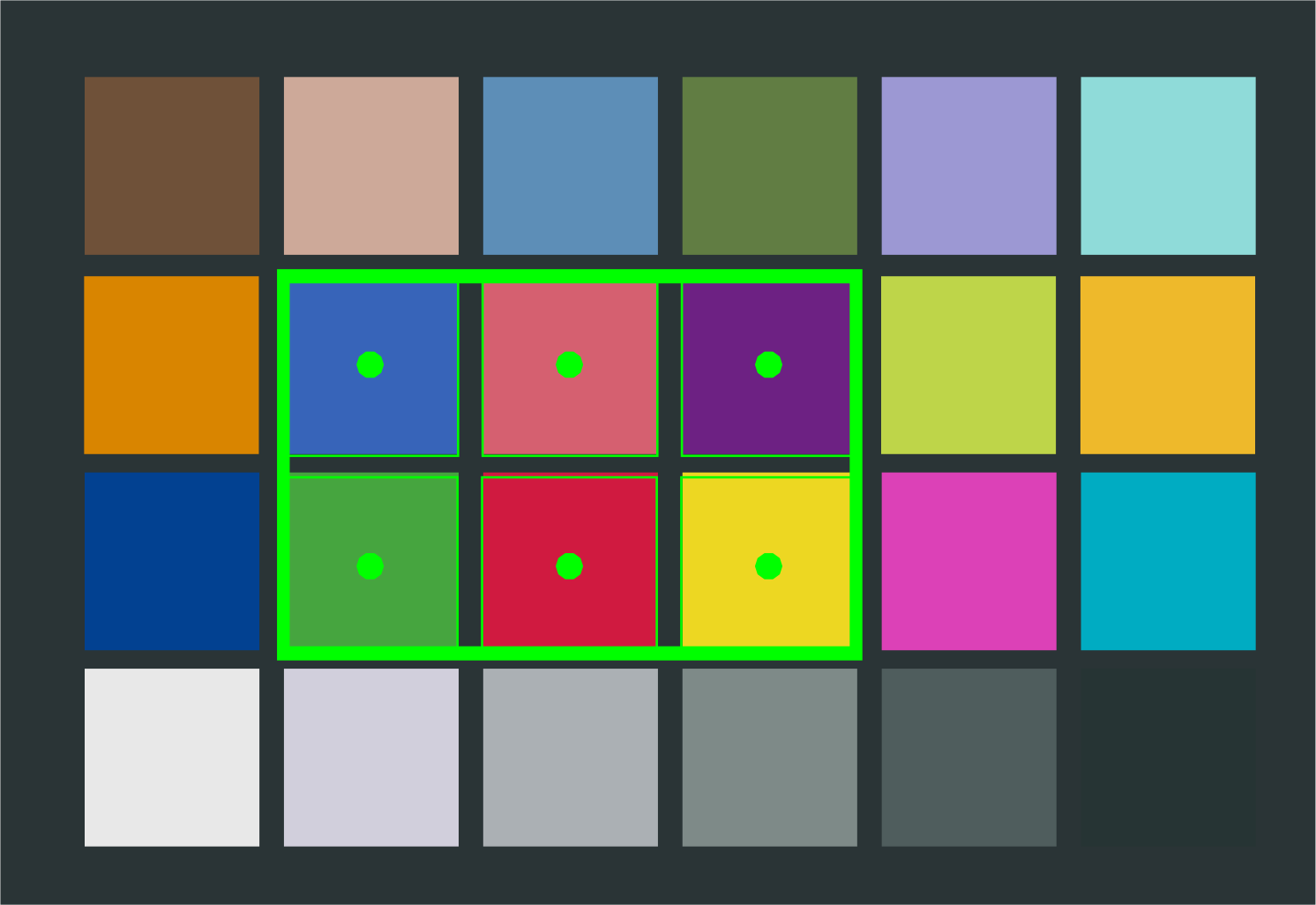} &
\includegraphics[width=1.30in]{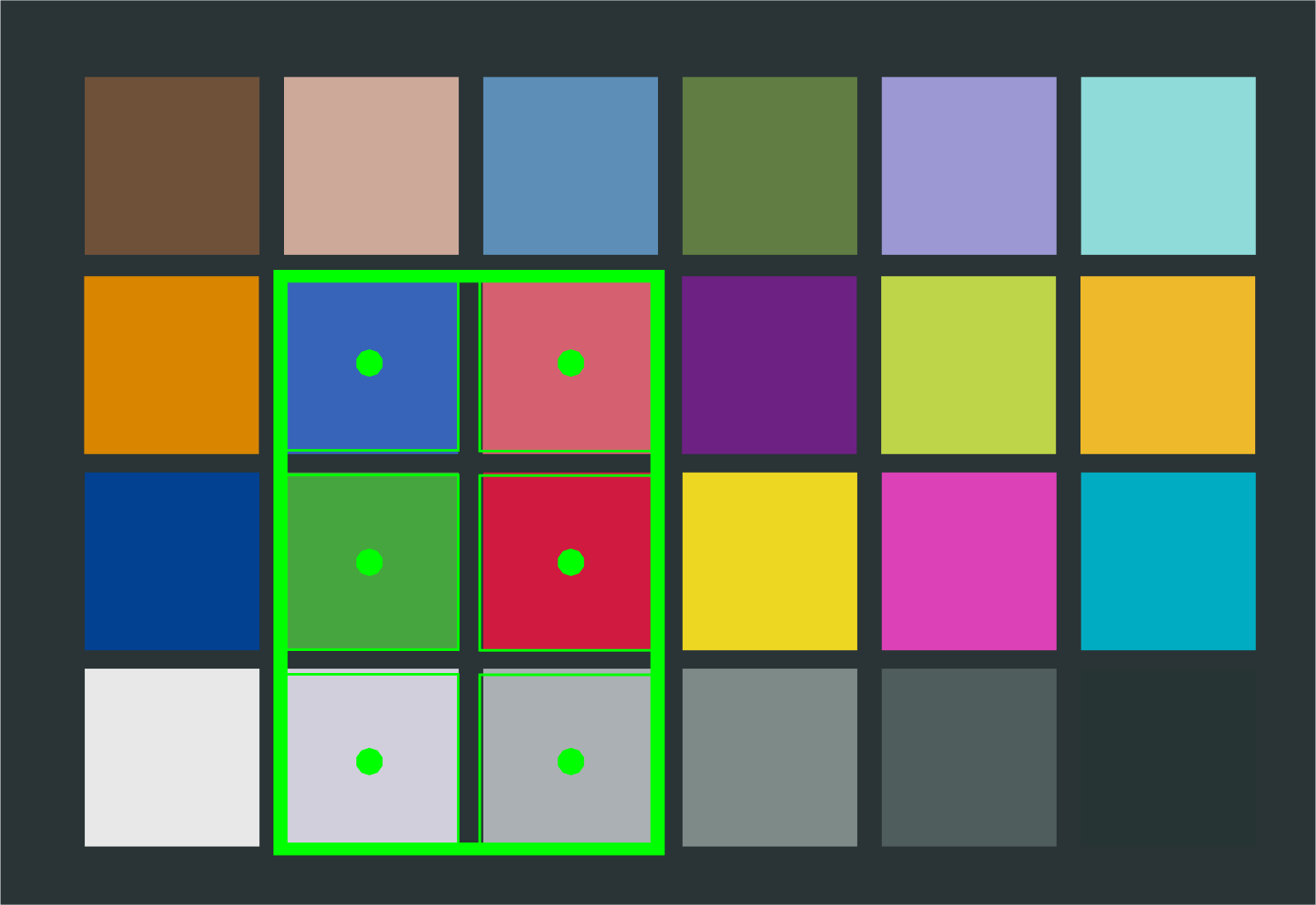} \\
$\delta=6$, $\theta=0$ &
$\delta=7$, $\theta=90$ \\
\end{tabular}
\caption{Examples for the estimation of the parameters $\theta$ and $\delta$ that represent, respectively, the rotation and the shift of the detected ColorChecker patches.}
\label{fig:optimization_proc}
%\vspace{-0.5cm}
\end{figure}

Then, the parameters $\theta$ and $\delta$ are estimated. They represent, respectively, the rotation and the shift of the detected ColorChecker patches with respect to the original CC model. For the parameters estimation, the following cost function is defined:

\begin{equation}
J(\theta,\delta) = || X^* - Rt(Shf(X_c, \delta), \theta) ||_2^2
\end{equation}

\noindent
where $Rt$ is a rotation operator and $Shf$ is a shift operation, $X_c$ is the detected ColorChecker patches subset, and $X^*$ is the original color model template CC. The parameters $\theta$ and $\delta$  that minimize this cost function is defined as (see \figurename~\ref{fig:optimization_proc}):

\begin{equation}
[\hat{\theta}, \hat{\delta}] = \arg \min_{\theta, \delta}  J(\theta ,\delta),
\end{equation}

\noindent
for angle $\theta \in \{0,90,180,270\}$ and step $\delta \in \{1,2, \cdots , n\}$.

An estimated position of the CC model is obtained with these parameters. The inverse of the homography matrix H is applied to obtain a model projection (hypothesis) in the image as shown in \figurename~\ref{fig:position_estimate}. For the validation of the obtained hypothesis  (11) a cost function similar to the one proposed by Ernst et al. \cite{Ernst2013CheckCalibration} is used:

\begin{equation}
E(p) = \sum_k \left(1 - \frac{ \mu_{k,p} r_k}{||\mu_{k,p}|| ||r_k||} \right) + \sum ||\sigma_{k,p}||^2,
\end{equation}

\noindent
where $\mu_{k,p}$ and $\sigma_{k,p}$ denotes the mean colors and standard deviations for each hypothesis $p$. The parameter $r_k$ is a reference color in the CC model. In this case, $k$ varies from 1 to 24 since the ColorChecker Classic has 24 patches.

% and $p$ is the index of each hypothesis

In this case, differently, from Ernst et al. \cite{Ernst2013CheckCalibration}, the cosine of the angle is used as a similarity function. This function is more robust to color changes than the Euclidean distance. 

Due to the clustering process (step 8), several hypotheses of the same object could be generated. The last step (13) aims at eliminating the redundant hypotheses and select those that represent a CC in the image. The redundant candidates are those that represent the same CC in the scene. To eliminate redundancy, the hypotheses that present an overlap are selected. For that, the intersection over union area of the bounding box of each candidate is used, and the lower cost, $E(p)$, hypotheses are selected. In general, the number of CCs present in the scene is known a priori. One of the advantages of this approach is that it can handle several CCs in the image. Assuming that $N$ ColorChecker were used in the scene, the selection of the hypotheses would be determined as follows: (1) select the N lower cost hypotheses; (2) hypotheses presenting a cost smaller than a threshold are selected.

\section{Experiments and Results}

In this section, the performance of the proposed methods are evaluated. The experiment is performed on a synthetic and a real ColorChecker dataset (GMCC) \cite{gehler2008bayesian}. The proposed method is referred to as MCCNetFind and a variation of this method, named MCCFind, which does not contain the localization step is also analyzed, showing the importance of the DetectNet in the detection process. 

The proposed method is compared with the CCFind and MacDuff, using the following metrics for the Intersection Over Union (IOU): True Positive (TP), False Positive (FP), False Negative (FN), Accuracy (Acc), Precision (Prec), Recall (Rec), F-Measure (F-Meas) and mean Average Precision (mAP). To analyze the quality of the results, three metrics based on IOU and cosine similarity are also used: 

$$a_0 = \frac{area(B_p \bigcap B_{gt})}{area(B_p \bigcup B_{gt})}$$
$$a_1 = \frac{1}{N}  \sum_i \frac{area(C_p^i \bigcap C_{gt}^i)}{area(C_p^i \bigcup C_{gt}^i)}$$
$$a_2 = \frac{1}{N} \sum_i \cos(\mu_{gt}^i, \mu_p^i) = \frac{1}{N} \sum_i \frac{ \mu_{gt}^i \mu_p^i}{||\mu_{gt}^i|| ||\mu_p^i||}$$

\noindent
where $N$ is the number of charts, $B_p$ is the predicted bounding box, $B_{gt}$ is the  ground truth bounding box, $C_p^i$ and $C_{gt}^i$ are predicted areas and ground truth areas of patches $i$ respectively, $\mu_p$ and $\mu_{gt}$ the mean color vector for each patch color. The performance measure $a_0$ is used to identify the correctly located CC color target. The metrics $a_1$ and $a_2$ define the recognition performance of all charts and its associated colored patches, respectively.

\subsection{Training the DetectNet}

\begin{figure}[!t]
\centering
\includegraphics[width=3.3in]{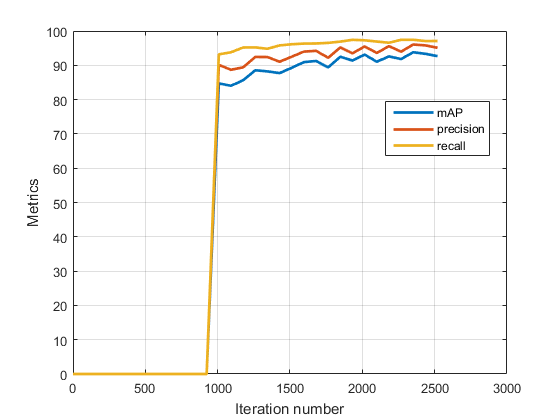}
\caption{Plot of the validation metrics, precision, recall and mean average precision vs the number of iterations of the DetectNet.}
\label{fig:net_metric}
\end{figure}

For the CCC localization, the DetectNet neural network using the Caffe framework\footnote{https://github.com/NVIDIA/caffe} was applied. We defined a new renderer layer to generate the synthetic data and the following hyperparameters were used: size of training set 5000; size of validation set 1000; image resolution $1024\times640$; learning rate $10^{-4}$; epochs 30; batch size 30; learning rate policy fixed; momentum 0.9; weight decay $10^{-6}$. \figurename~\ref{fig:net_metric} shows the evaluation metrics of the training process. The mAP, precision and recall values obtained for the trained model on the validation set was 92.67\%, 95.11\% and 97.09\% respectively.

% \begin{figure}[!t]
% \centering
% \includegraphics[width=2.75in]{netloss.eps}
% \caption{Plot of the loss vs the number of iterations of the DetectNet.}
% \label{fig:net_loss}
% \end{figure}

\subsection{{Experiment using synthetic images}}
\subsubsection{Protocol}

The experiments are performed as follows: Number of instances (images) $N=1000$; Rigid transformation parameters, the rotation matrix $Rt=[r_x r_y r_z]$ where $r \in [-\pi/2,\pi/2]$; and the translation matrix $Tr=[t_x t_y t_z]$ where $t \in [-10,-30]$; Background (BG) image was obtained from the gratisography website\footnote{https://gratisography.com/}; Number of the ColorChecker models $M= \{1,2,3,4,5\}$.

\begin{figure}[!t]
\centering
\begin{tabular}{c}
\includegraphics[width=3.0in]{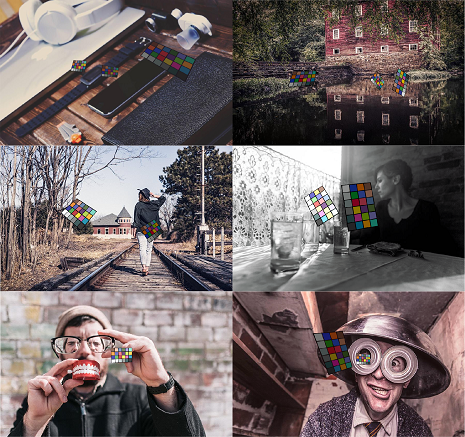} 
\end{tabular}
\caption{Examples of images from the synthetic dataset with single and multiple ColorChecker.}
\label{fig:synthetic_dataset}
\end{figure}

For each iteration, a new position of the ColorChecker model is obtained randomly (this is the matrix $[Rt | Tr]$) and projected over a randomly selected BG image. \figurename~\ref{fig:synthetic_dataset} depicts some of the $1000$ generated images used for this experiment.

\subsubsection{Results}

Table \ref{tab:synthetic_datasets} displays the results obtained using the synthetic database. For each method, all images were visually analyzed and the following measures: TP, FP, FN, Acc, Prec, Rec and F-Measure, were calculated. As can be seen, the proposed MCCNetFind method presents a precision of $0.97$, which indicates that the method is successful for automatic ColorChecker detection. In all cases, the proposed method exhibits an accuracy of $0.85$ and an F-Measure of $0.92$ which are better results compared to the results by CCFind and MacDuff. The synthetic dataset presents very complex examples with changes in position and color of the CC, for which the MCCNetFind method presented a much better recall values compared to the other methods.

\begin{table}[ht]
\centering
\caption{Results for the synthetic dataset with a single ColorChecker.}
\label{tab:synthetic_datasets}
\scalebox{0.85}{
\begin{tabular}{c}
\begin{tabular}{lllllllll}
\hline
Methods&     TP&  FP&   FN&  Total&   Acc&  Prec&   Rec& F-Meas \\
\hline
    CCFind& 334& 142&  524&   1000&  0.33&  0.70&  0.39&   0.50 \\
   MacDuff&  29& 199&  772&   1000&  0.03&  0.13&  0.04&   0.06 \\
   MCCFind& 536&   3&  461&   1000&  0.54&  \textbf{0.99}&  0.54&   0.70 \\
MCCNetFind& 855&  29&  116&   1000&  \textbf{0.85}&  0.97&  \textbf{0.88}&  \textbf{0.92} \\ 
\hline
\end{tabular}\\\\
\multicolumn{1}{p{3.0in}}{TP: true positive, FP: false positive, FN: false negative, Acc: accuracy, Prec: precision, Rec: recall, F-Meas: f-measure.}
\end{tabular}
}
\end{table}

\begin{figure}[!t]
\centering
\includegraphics[width=3.15in]{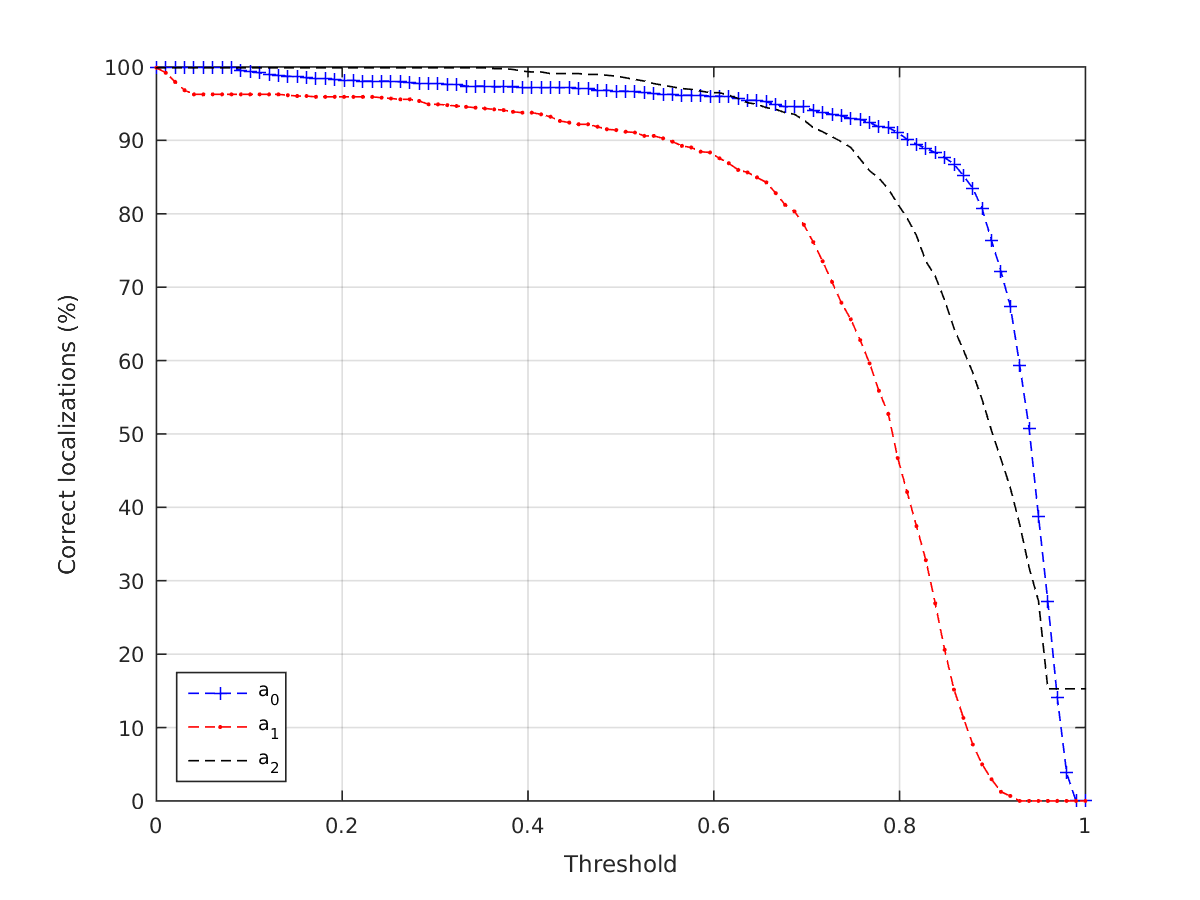}
\caption{Plot of the accuracy of detected CC (TP+FP) against $a_0$, $a_1$ and $a_2$ values for the MCCNetFind method.}
\label{fig:synthetic_metric}
%\vspace{-0.5cm}
\end{figure}

The $a_0$, $a_1$ and $a_2$ metrics were calculated to measure the quality of MCCNetFind method (only for the CCs detected, TP and FP). \figurename~\ref{fig:synthetic_metric} shows the localization accuracy of the detected CC as a function of the correct localization. The $a_1$ metric value shows that more than 70\% of detected ColorCheckers were obtained for values smaller than 0.75 and in the case of $a_0$ and $a_2$ more than 95\%. Therefore, it is clear that not only the method is accurate, but the results show a high quality standard.

The experiments reveals that the proposed method can perform detection of multiple CC in an image, while both CCFind and MacDuff are unable to detect multiple ColorChecker's. The Table \ref{tab:mult_synthetic_datasets} presents the results of the method using a synthetic database with multiple ColorChecker. In the MCCFind case, it is necessary to know a priori the number of ColorChecker's in the scene.

\begin{table}[ht]
\centering
\caption{Results for the synthetic dataset with multiple ColorChecker.}
\label{tab:mult_synthetic_datasets}
\scalebox{0.85}{
\begin{tabular}{c}
\begin{tabular}{lllllllll}
\hline
Methods&      TP&  FP&   FN&  Total&   Acc&  Prec&   Rec& F-Meas \\
\hline
   MCCFind& 1287&  11& 1154&   2452&  0.52&  \textbf{0.99}&  0.53&   0.69 \\
MCCNetFind& 2039& 122&  291&   2452& \textbf{0.83}&  0.94&  \textbf{0.88}&   \textbf{0.91} \\
\hline
\end{tabular}\\\\
\multicolumn{1}{p{3.0in}}{TP: true positive, FP: false positive, FN: false negative, Acc: accuracy, Prec: precision, Rec: recall, F-Meas: f-measure.}
\end{tabular}
}
\end{table}

\begin{figure}[!t]
\centering
\begin{tabular}{c}
\includegraphics[width=3.0in]{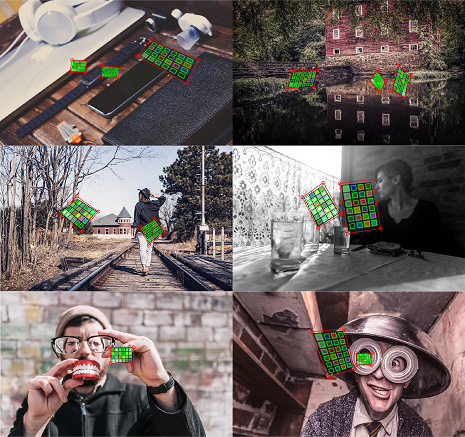} 
\end{tabular}
\caption{Results of the MCCNetFind method for the synthetic dataset.}
\label{fig:mult_synthetic_dataset}
\end{figure}

\figurename~ \ref{fig:mult_synthetic_dataset} illustrates examples of the results in images of multiple CCCs. A desirable feature of this method is the high processing speed since most of the methods described in the literature  are slow. The algorithms implementation was done using Matlab and C++/Python. The experiments were conducted on an Intel \textregistered Core$^{TM}$ i7 with 2.8 GHz and a Nvidia GeForce GTX 980 Ti. The system required on average $0.89 \pm 0.43$ seconds per image to detect the CCC in the test sequence with the C++/Python implementation. The C++/Python version\footnote{\url{https://github.com/pedrodiamel/colorchacker-detection}} was made available in a repository. 

\subsection{Experiment using real images}
\subsubsection{Protocol}

\begin{figure}[!t]
\centering
\includegraphics[width=2.75in]{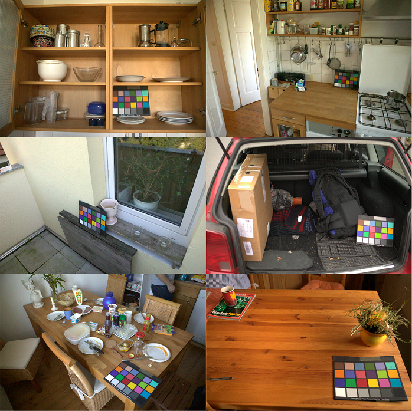}
\caption{Examples of images from the GMCC dataset that contain images with real ColorChecker's.}
\label{fig:gmcc}
\end{figure}

This experiment was performed on the GMCC dataset\cite{gehler2008bayesian}. The images were captured using a high-quality digital SLR camera in RAW format, as shown in \figurename~\ref{fig:gmcc}, so it is free from any color correction. Using the freely available software dcRAW\footnote{\url{http://www.cybercom.net/~dcoffin/dcraw/}} the images were demosaiced and converted into uncompressed linear 16-bit files. This process was done with particular attention to converting the images using always the same multiplicative gains to bypass the camera Automatic White Balance (AWB) estimation. The dataset consists of 569 images.

\subsubsection{Results}

\begin{figure}[!t]
\centering
\includegraphics[width=2.75in]{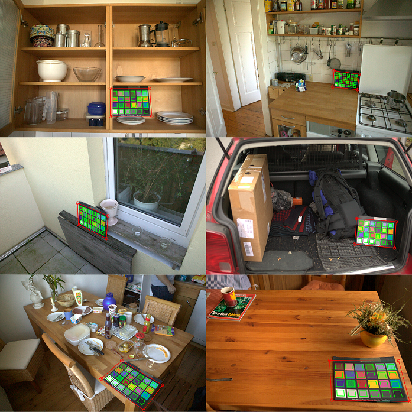}
\caption{Results of the MCCNetFind method for the GMCC dataset.}
\label{fig:resultsmccnet}
\end{figure}

Table \ref{tab:real_datasets} shows the results obtained for the GMCC dataset. The precision values for the proposed methods is maintained at $0.990$. The ColorChecker in real images usually presents less complex transformations than in the synthetic images, so the detection method most of the time shows better results with improved recall rate. In this case, the results of the proposed MCCFind method present an accuracy rate of $0.920$ and  F-Measure of $0.960$, showing that it improves the performance compared to the other methods. The proposed MCCNetFind method also outperforms (0.972 accuracy) previous algorithms from the Table \ref{tab:real_datasets} and obtains an F-Measure of $0.986$ with very high precision confirming its capability for automatic ColorChecker detection. \figurename~\ref{fig:resultsmccnet} portrays the MCCNetFind method's results for samples from the GMCC dataset.

\begin{table}[ht]
\centering
\caption{Results for the GMCC Dataset.}
\label{tab:real_datasets}
\scalebox{0.85}{
\begin{tabular}{c}
\begin{tabular}{lllllllll}
\hline
Methods&      TP&     FP&  FN&   Total&     Acc&    Prec&  Rec&    F-Meas \\
\hline
Kordecki \cite{Kordecki2014}
          &  440&    110&  19&     569&    0.770&    0.800& 0.960&      0.870 \\
    X-Rite&  306&    241&  22&     569&    0.540&    0.560& 0.930&      0.700 \\
    CCFind&  430&     36& 103&     569&    0.760&    0.920& 0.810&      0.860 \\
   MacDuff&   41&    180& 348&     569&    0.070&    0.190& 0.110&      0.130 \\
   MCCFind&	 523&      3&  43&     569&    0.920&    0.990& 0.920&      0.960 \\
MCCNetFind&  553&      3&  13&     569&    \textbf{0.972}&    \textbf{0.995}& \textbf{0.977}&      \textbf{0.986} \\
\hline
\end{tabular}\\\\
\multicolumn{1}{p{3.0in}}{TP: true positive, FP: false positive, FN: false negative, Acc: accuracy, Prec: precision,  Rec: Recall, F-Meas: F-Measure.}
\end{tabular}
}
\end{table}

\begin{figure}[t]
\centering
\begin{tabular}{cc}
\includegraphics[width=1.5in]{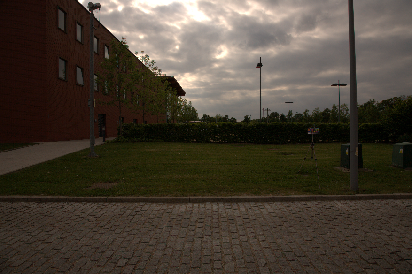} &
\includegraphics[width=1.5in]{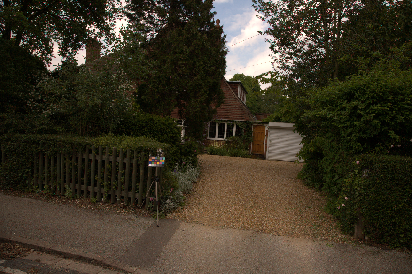} \\
\includegraphics[width=1.5in]{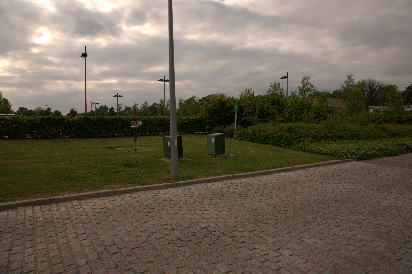} &
\includegraphics[width=1.5in]{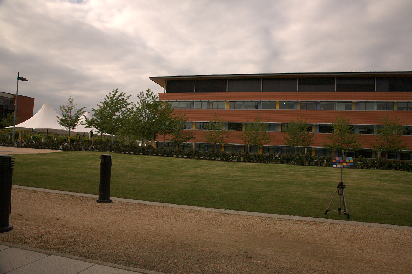} \\
\end{tabular}
\caption{Examples of images that present errors in the localization.}
\label{fig:error_detection}
\end{figure}

\begin{figure}[!t]
\centering
\includegraphics[width=2.75in]{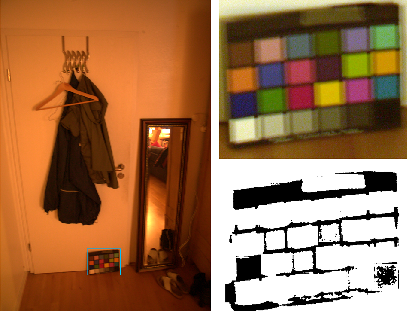}
\caption{Examples of images that present errors in the recognition.}
\label{fig:error_recognition}
\end{figure}

Most errors in the localization step (13 errors out of the 16) occurs where the ColorChecker's are too far away from the camera (see \figurename~\ref{fig:error_detection}). The neural network does not manage to generalize these cases because the render layer generates the patterns in a defined range distance from the camera in which examples of this type do not appear. The errors by the recognition system (3 errors) are due to blurring in ColorChecker's that makes it difficult the segmentation of the charts (see \figurename~\ref{fig:error_recognition}).

In this work, we demonstrate the feasibility of the use of synthetic images to train convolutional neural networks to detect ColorChecker's. Future works will be aimed at the creation of an end-to-end model based on deep convolutional networks for the task of detection, color patch recognition and estimation of the pose of multiple ColorChecke's types.

\section{Conclusion}

We presented a deep learning based ColorChecker detection method that showed a high accuracy and precision rates. The proposed solution is fast and completely automatic. Also, an algorithm to find the checker minimum enclosing as well as a variation of the clustering algorithm HCA \cite{gil2006general} ensuring that the method can detect multiple CC were presented. A synthetic dataset was generated to evaluate the results. The proposed method showed an accuracy improvement of over 0.20 other methods in the state-of-the-art and high precision and recall rates. The GMCC dataset was used to evaluate real images, and the obtained accuracy was 0.972, demonstrating a significant increase compared to other methods in the literature. We also tested the influence of the deep learning detection step by applying the recognition method directly in the image. Results show that the deep learning detection improves the accuracy in 4\% in the GMCC dataset, but for the synthetic dataset, the improvement was of 31\%.

\section*{Acknowledgments}
This work was supported by the research cooperation project between Motorola Mobility LLC (a Lenovo Company) and CIn-UFPE. Tsang Ing Ren, Pedro D. Marrero Fernandez and Fidel A. Guerrero-Pe\~{n}a gratefully acknowledge financial support from the Brazilian government agency FACEPE. The authors would also like to thank Leonardo Coutinho de Mendon\c{c}a, Alexandre Cabral Mota, Rudi Minghim and Gabriel Humpire for valuable discussions.

%\clearpage
%\printglossaries

%
\bibliographystyle{IEEEbib}
%\footnotesize
\bibliography{refs}

\end{document}